\documentclass[sigconf, nonacm]{acmart}
\renewcommand\footnotetextcopyrightpermission[1]{}
\AtBeginDocument{%
  }

\usepackage{float}
\usepackage{subcaption}
\usepackage{graphicx}

\begin{document}

\title{Glove2UAV: A Wearable IMU-Based Glove for Intuitive Control of UAV}


\author{Amir Habel*}
\orcid{0009-0009-4239-219X}

\affiliation{%
  \institution{Skolkovo Institute of Science and Technology}
  \city{Moscow}
  \country{Russia}
}
\email{amir.habel@skoltech.ru}

\author{Ivan Snegirev*}

\affiliation{%
  \institution{Skolkovo Institute of Science and Technology}
  \city{Moscow}
  \country{Russia}
}
\email{ivan.snegirev@skoltech.ru}

\author{Elizaveta Semenyakina*}
\affiliation{%
  \institution{Skolkovo Institute of Science and Technology}
  \city{Moscow}
  \country{Russia}
}
\email{elizaveta.semenyakina@skoltech.ru}

\author{Miguel Altamirano Cabrera}
\orcid{0000-0002-5974-9257}
\affiliation{%
  \institution{Skolkovo Institute of Science and Technology}
  \city{Moscow}
  \country{Russia}
}
\email{m.altamirano@skoltech.ru}

\author{Jeffrin Sam}
\affiliation{%
  \institution{Skolkovo Institute of Science and Technology}
  \city{Moscow}
  \country{Russia}
}
\email{jeffrin.sam@skoltech.ru}

\author{Fawad Mehboob}
\affiliation{%
  \institution{Skolkovo Institute of Science and Technology}
  \city{Moscow}
  \country{Russia}}
\email{Fawad.Mehboob@skoltech.ru}

\author{Roohan Ahmed Khan}
\affiliation{%
  \institution{Skolkovo Institute of Science and Technology}
  \city{Moscow}
  \country{Russia}
}
\email{roohan.khan@skoltech.ru}

\author{Muhammad Ahsan Mustafa}
\affiliation{%
  \institution{Skolkovo Institute of Science and Technology}
  \city{Moscow}
  \country{Russia}}
\email{Ahsan.Mustafa@skoltech.ru}

\author{Dzmitry Tsetserukou}
\orcid{0000-0001-8055-5345}
\affiliation{%
  \institution{Skolkovo Institute of Science and Technology}
  \city{Moscow}
  \country{Russia}}
\email{d.tsetserukou@skoltech.ru}

\authornote{These authors contributed equally to this research.}

\renewcommand{\shortauthors}{A. A. Habel et al.}


\begin{abstract}
This paper presents Glove2UAV, a wearable IMU-glove interface for intuitive UAV control through hand and finger gestures, augmented with vibrotactile warnings for exceeding predefined speed thresholds. To promote safer and more predictable interaction in dynamic flight, Glove2UAV is designed as a lightweight and easily deployable wearable interface intended for real-time operation. Glove2UAV streams inertial measurements in real time and estimates palm and finger orientations using a compact processing pipeline that combines median-based outlier suppression with Madgwick-based orientation estimation. The resulting motion estimations are mapped to a small set of control primitives for directional flight (forward/backward and lateral motion) and, when supported by the platform, to object-interaction commands. Vibrotactile feedback is triggered when flight speed exceeds predefined threshold values, providing an additional alert channel during operation. We validate real-time feasibility by synchronizing glove signals with UAV telemetry in both simulation and real-world flights. The results show fast gesture-based command execution, stable coupling between gesture dynamics and platform motion, correct operation of the core command set in our trials, and timely delivery of vibratile warning cues.


\end{abstract}


\maketitle

\begin{figure}[!t]
    \centering
    \includegraphics[width=0.9\linewidth]{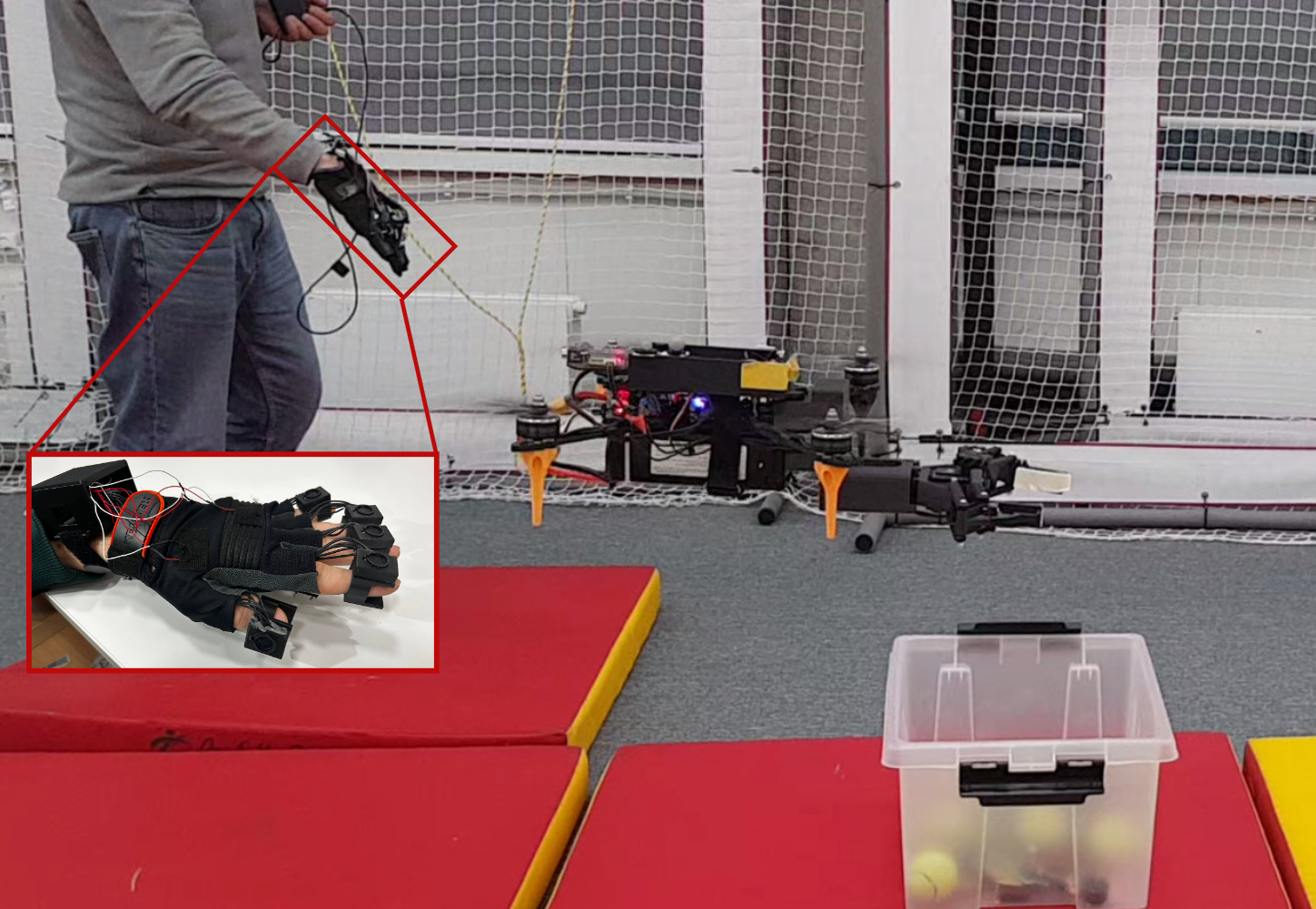}
    \caption{Overview of the proposed IMU-glove UAV teleoperation system.}
    \label{fig:related_drone_interfaces}
\end{figure}

\section{Introduction}

As robotic systems are increasingly deployed across work and everyday scenarios, the demand for intuitive and effective robot control interfaces continues to grow. Wearable interfaces \cite{DelPreto20PlugAndPlay} leverage natural human motion and enable intuitive interaction with low deployment overhead, making them attractive where rapid training and operator mobility are important.

Unmanned aerial vehicles (UAVs) are among the most demanding mobile robotic platforms: flight dynamics and sensitivity to control inputs make operation precision-critical, and operator errors can be particularly consequential. With expanding UAV applications, there is a growing need for control interfaces that remain understandable and reliable even for users with limited piloting experience \cite{Tezza19HumanDroneSurvey}.

Existing approaches range from conventional two-stick transmitters to immersive and eye-tracking interfaces \cite{GazeRace}. AeroVR \cite{Yashin19AeroVR} explores VR teleoperation with tactile feedback for aerial manipulation, using immersive visual context and vibrotactile cues for grasp confirmation. DroneRacing \cite{Pfeiffer21DroneRacing} and OmniRace \cite{Serpiva24OmniRace} demonstrates vision-based low-level gesture control of a racing drone via 6D hand pose estimation and gesture recognition, targeting highly dynamic scenarios. A handheld motion-controller approach \cite{Trinitatova25Handheld} provides an alternative to traditional transmitters, emphasizing one-handed operation and broad platform compatibility.

Against this background, we investigate an IMU-glove for UAV control augmented with tactile cues intended to support operator awareness and confidence during dynamic flight. Unlike VR approaches that require specialized hardware, the proposed interface preserves expressive gesture control with substantially lower deployment overhead. Compared to vision-based interfaces that can be sensitive to lighting, camera placement, and environmental constraints, an IMU glove provides a stable input channel that does not depend on external visual tracking or a specific spatial setup \cite{Gromov19PointingIMU, Harrison25IMUSmartGlove, Lin18}. Relative to handheld motion controllers, the wearable form factor enables palm- and finger-level control while preserving user mobility and freedom of movement.

Overall, our approach combines mobility, minimal external infrastructure, a compact set of control primitives, and an integrated tactile warning channel, making it promising for scenarios beyond laboratory settings.

\section{System Architecture}

\begin{figure}[htbp]
    \centering
    \includegraphics[width=1\linewidth]{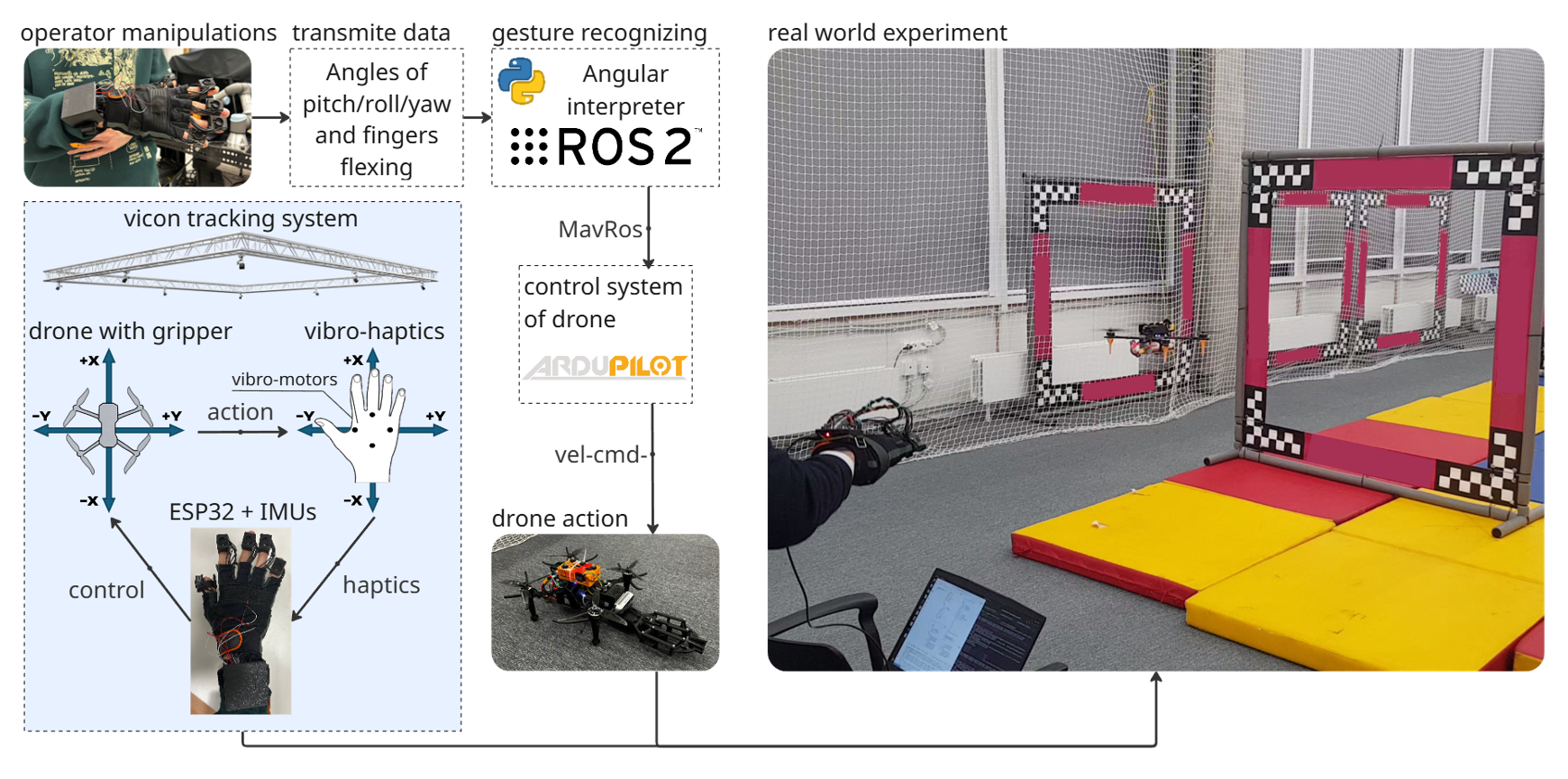}
    \caption{System architecture.}
    \label{fig:system_arch}
\end{figure}

The system implements a wearable gesture-based interface for UAV control with an integrated vibrotactile feedback channel. The architecture comprises four components: the operator, the IMU glove, the PC-based processing module, and the UAV (Fig.~\ref{fig:system_arch}). The glove captures palm and finger motions and streams inertial data to the PC, where signals are filtered and mapped to control commands that are transmitted to the UAV. In the reverse direction, UAV telemetry is used to generate vibrotactile warnings that complement the operator’s visual monitoring. Overall, the loop includes three data flows: (i) glove-to-PC IMU measurements , (ii) PC-to-UAV control commands, and (iii) UAV-to-PC telemetry for tactile cue generation.

\begin{figure} [t]
    \centering
    \includegraphics[width=0.95\linewidth]{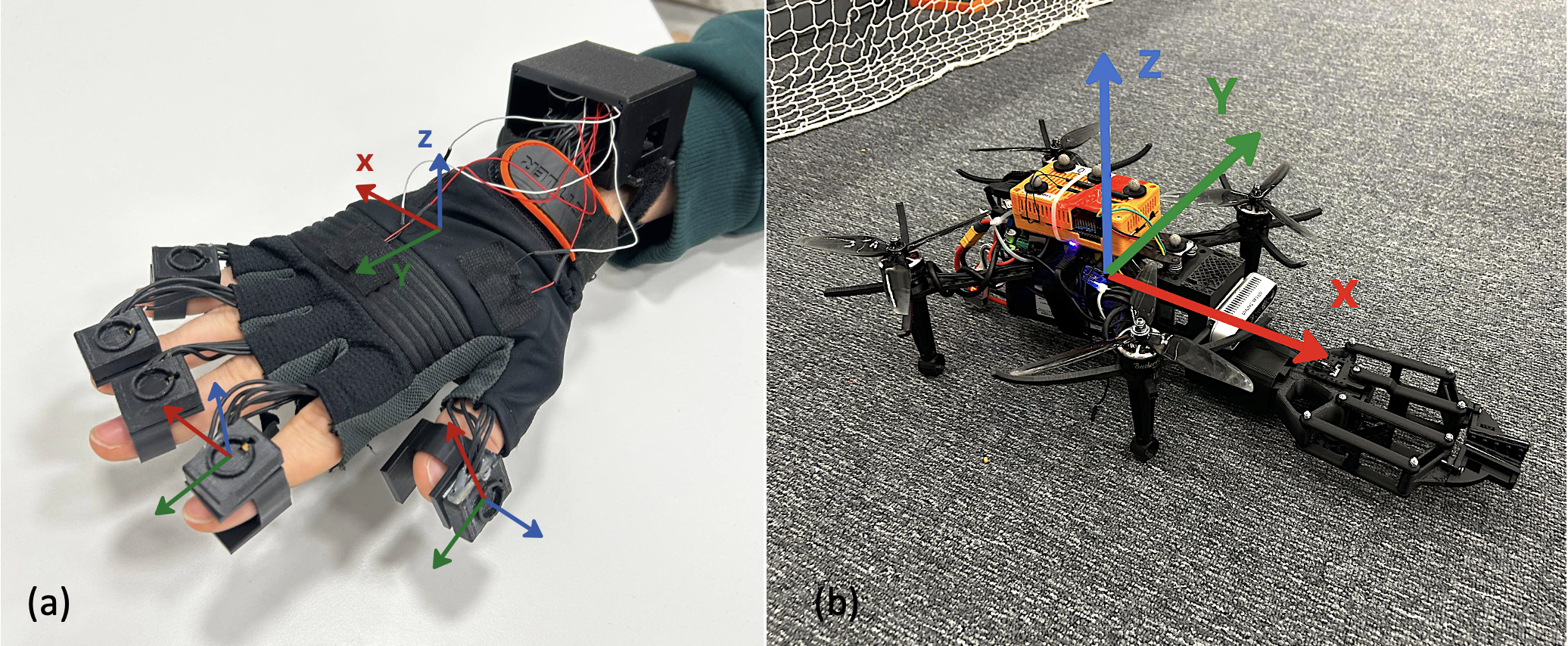}
    \caption{ (a) Wearable IMU-glove with integrated vibrotactile
actuators. (b) UAV platform used for validating the proposed interface.}
    \label{fig:imu_glove}
\end{figure}

The glove hardware includes IMU sensors on the palm and fingers, an ESP32 microcontroller, vibrotactile actuators, and a power module (Fig.~\ref{fig:imu_glove}(a)). The palm IMU serves as a reference for overall hand motion, while finger IMUs capture relative movements for gesture commands. The ESP32 performs synchronized sampling and wireless transmission to the processing module. Vibrotactile sensors notify the operator when potentially risky flight regimes are detected.

The UAV platform executes high-level motion commands via its onboard controller, while stabilization and low-level control remain within the autopilot loop (Fig.~\ref{fig:imu_glove}(b)). Flight-state data, including velocity, are used by the processing module to trigger speed-threshold vibrotactile warnings during operation.

\section{Gesture Processing and Control Logic}

\label{sec:signal-processing}

\begin{figure}[H]
    \centering
    \includegraphics[width=0.95\columnwidth]{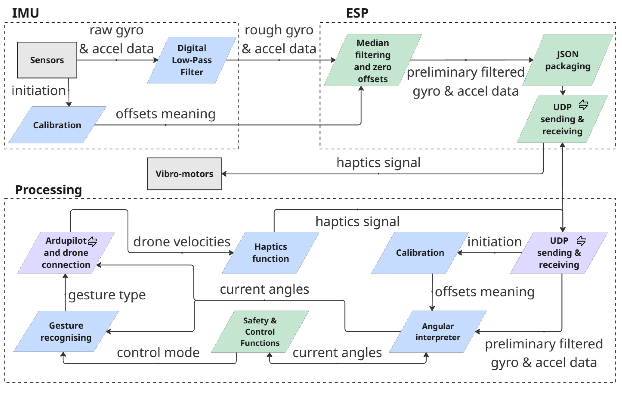}
    \caption{Unified IMU-glove processing and control pipeline for UAV teleoperation.}
    \label{fig:Scheme3).}
\end{figure}

The unified processing and control pipeline of the proposed IMU-glove interface is presented in Fig.~\ref{fig:Scheme3).}. Raw gyroscope and Accelerometer measurements are first preprocessed on board. The GY-91 modules provide built-in digital low-pass filtering and initial calibration. On the ESP32, we additionally suppress outliers using a sliding-window median filter:
\begin{equation}
y_k = \mathrm{med}(x_{k-n}, x_{k-n+1}, \ldots, x_k, \ldots, x_{k+n-1}, x_{k+n}),
\label{eq:median}
\end{equation}
followed by zero-offset compensation. The preliminary filtered stream is packaged into JSON and transmitted over Wi-Fi via UDP with a timestamp for reliable parsing on the PC.

After UDP reception, the main orientation estimation stage is executed.
For the finger IMUs, where motion is represented in a simplified 2D
setting \cite{Yang20}, we employ a first-order complementary filter. Accelerometer-based
tilt angles are computed as:
\begin{equation}
\boldsymbol{\theta}_a =
\begin{bmatrix}
\theta_x\\
\theta_y\\
\theta_z
\end{bmatrix}
=
\begin{bmatrix}
\arctan2(a_y,a_z)\\
\arctan2(-a_x,\sqrt{a_y^2+a_z^2})\\
0
\end{bmatrix},
\label{eq:acc_tilt}
\end{equation}
where only roll $\theta_x$ and pitch $\theta_y$ are estimated, while
$\theta_z$ is set to zero. The gyroscope-based estimate is obtained by
numerical integration:
\begin{equation}
\theta_{\omega,t} = \theta_{t-1} + \omega_x \Delta t,
\label{eq:gyro_int}
\end{equation}
and the fused angle is computed as:
\begin{equation}
\hat{\theta}_t = \alpha \theta_{\omega,t} + (1-\alpha)\theta_{a,t},
\quad \alpha \in [0,1].
\label{eq:comp}
\end{equation}

For the wrist IMU, we employ the Madgwick filter \cite{Madgwick10} to estimate roll, pitch, and yaw. In parallel, the Reset pose, Zero-back alignment, and Finger Lock procedures are applied to stabilize the zero reference pose and reduce ambiguity in gesture recognition. The Zero-back alignment mechanism continuously corrects the reference pose by tracking deviations of the current orientation from the initial zero state, thereby compensating for accumulated errors and drift. The Finger Lock procedure temporarily disables gesture recognition during active UAV control, preventing false activations caused by unintentional finger movements. In addition, a manual reference pose reset (Reset pose) is provided, increasing the system’s robustness to unexpected failures and initialization errors.

Gesture commands have been explored for controlling high-level UAV behavior \cite{Akagi19HighLevelUAVGestures}. Based on the estimated orientations, the system implements three gesture-based control functions: gripper open/close, step-based altitude control, and continuous wrist attitude control (roll/pitch/yaw). The recognized gesture type and wrist orientation values are forwarded to a ROS-based drone connection module for command delivery in both simulation and real-world flights.

\section{Experimental Evaluation}

\begin{figure*}[t]
    \centering
    \includegraphics[width=\textwidth]{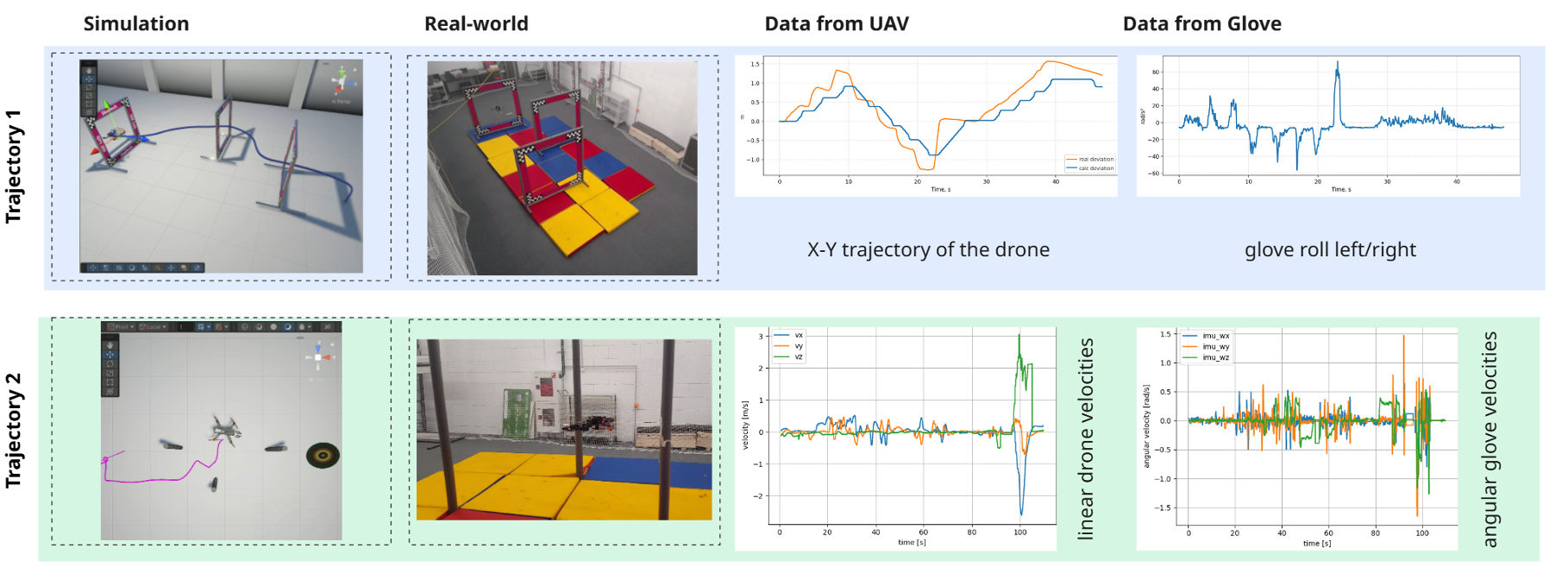}
    \caption{Experiment 1 (trajectory-following control): comparison of simulation and real-world flights for two representative trajectories.}
    \label{fig:system-overview}
\end{figure*}

\begin{figure*}[t]
    \centering
    \includegraphics[width=0.9\textwidth]{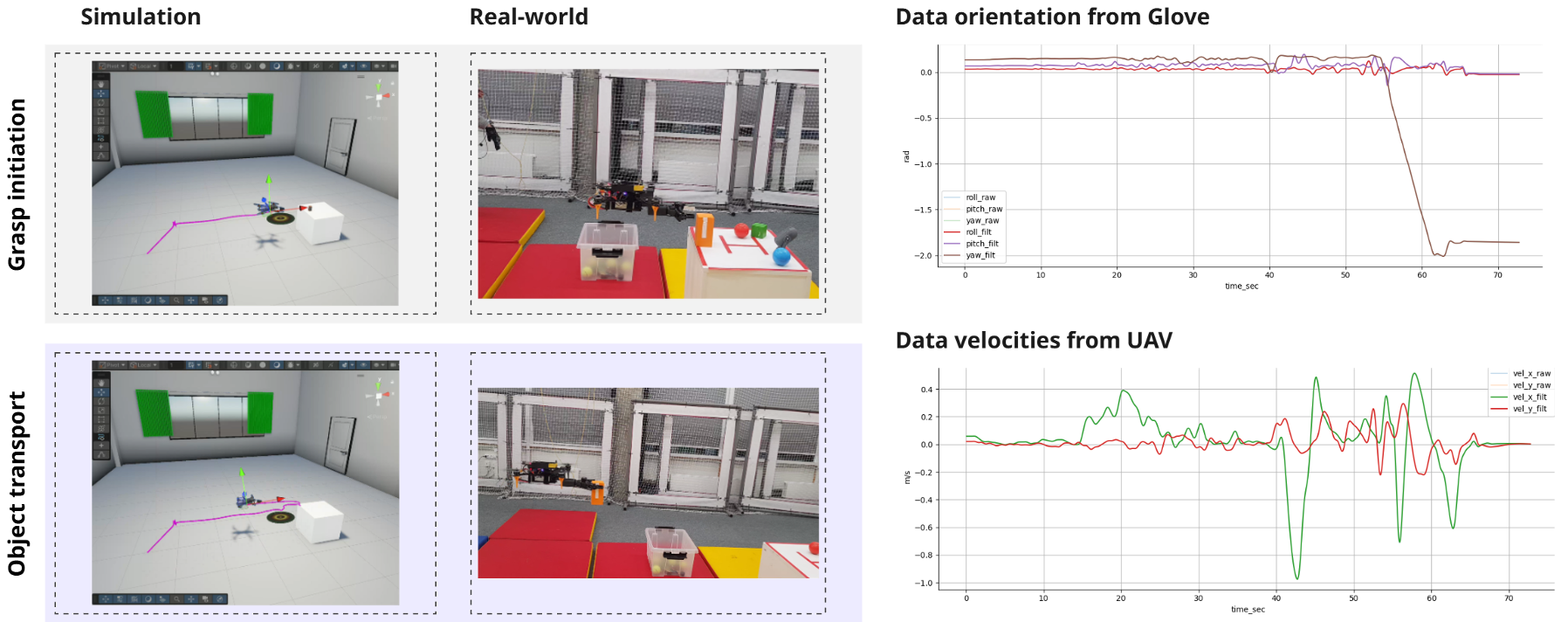}
    \caption{Experiment 2 (object grasp). Comparison of simulation and real-world execution using the proposed interface.}
    \label{fig:experiment2}
\end{figure*}

This section presents the results of an initial validation of the proposed wearable UAV control interface. The experiments were aimed at (1) verifying the consistency between gesture commands extracted from the IMU-glove data and UAV motion dynamics, and (2) demonstrating the operability of the command set in flight scenarios and during object interaction. The evaluation was based on UAV telemetry and the glove’s inertial measurement stream processed in real time. This experimental assessment serves as a feasibility validation of the gesture–command loop rather than a detailed quantitative accuracy analysis, which is left for future work.

Consistency between gesture kinematics and UAV telemetry was assessed by comparing the UAV linear velocities with the corresponding angular velocities obtained from the IMU glove. The synchronized time-series plots show the expected coherent dynamics: changes in gesture commands are accompanied by corresponding changes in telemetry, supporting the selected logic for mapping palm and finger movements to control commands.

\subsection{Experiment 1: Trajectory Flight Control.}

In the first set of experiments, the operator flew the UAV along a predefined trajectory using the IMU glove. Control was implemented by mapping hand orientation to directional motion commands: pitch motion of the hand commanded UAV forward/backward movement, while right/left roll commanded lateral motion to the right/left. The operator’s task was to guide the UAV from the start configuration to the finish while avoiding obstacles using these gesture commands. A 6-DoF UAV was used in this experiment. All trials were performed by a single operator and repeated five times to assess consistency. One run was defined as a complete flight along the prescribed trajectory with obstacle avoidance. In all repetitions, IMU-glove data and UAV telemetry were recorded and synchronized in real time. Fig.~\ref{fig:system-overview} presents the results of Experiment 1 for two representative trajectories: the Real-world column shows a photograph of the corresponding real run, while the Simulation column shows a trajectory visualization replayed in a simulated environment based on telemetry from that real flight. The right-hand side shows synchronized time series of UAV telemetry (Data from UAV) and IMU-glove measurements (Data from Glove) used to compare gesture dynamics with platform motion. Overall, the results confirm the operability of the continuous control loop and a consistent platform response to changes in gesture commands.

\subsection{Experiment 2: Object Interaction}

In the second set of experiments, we evaluated finger gestures intended for object-interaction commands when supported by the platform. The grasp command was generated by a finger-closing gesture: when the operator closed their fingers, the gripper closed and grasped the object. In this experiment, the UAV had 7 DoF due to an additional gripper for object manipulation. All trials were performed by a single operator and repeated five times. One run consisted of approaching the object, grasping it, and departing the interaction area. Fig.~\ref{fig:experiment2} shows key phases of the Experiment 2 scenario (grasp initiation and object transport): for each phase, the Real-world column provides a photograph of the real execution, while the Simulation column shows a trajectory visualization replayed in a simulated environment from the telemetry of the corresponding real run. The right-hand side presents synchronized time series of IMU-glove data (Data from Glove) and UAV telemetry (Data from UAV). The observed sequence confirms that finger gestures can serve as a separate informative control channel within the demonstrated scenario, complementing palm-motion-based flight control.

The results confirm the feasibility of gesture-primitive-based control for directional flight and object interaction. Vibrotactile warnings were triggered when flight speed exceeded predefined thresholds and qualitatively encouraged more cautious behavior (reduced gesture intensity and fewer abrupt maneuvers), yielding lower speeds and smoother motion. A quantitative evaluation of vibrotactile feedback is left for future work.
The wearable form factor reduces reliance on external infrastructure and may lower the entry barrier for UAV operation, while vibrotactile feedback may support operator confidence and perceived safety in dynamic flight regimes \cite{Labazanova19SwarmGlove}.
\section{Conclusion}

This work presents a wearable IMU-glove for gesture-based UAV control with vibrotactile feedback. Simulation and real-world flights demonstrate feasibility, with prompt gesture-to-command execution, consistent coupling between control primitives and UAV motion, and timely warning cues.

The system uses a lightweight real-time pipeline with median-based outlier suppression, Madgwick orientation estimation, and complementary filtering in a compact, deployable form.
Future work will expand user studies, benchmark against alternative UAV interfaces, test transferability across UAV platforms and settings, and explore deep-learning–based gesture recognition \cite{Jaramillo22}.

\section*{Acknowledgements} 
Research reported in this publication was financially supported by the RSF grant No. 24-41-02039.

\nocite{*}


\begin{thebibliography}{16}


\ifx \showCODEN    \undefined \def \showCODEN     #1{\unskip}     \fi
\ifx \showISBNx    \undefined \def \showISBNx     #1{\unskip}     \fi
\ifx \showISBNxiii \undefined \def \showISBNxiii  #1{\unskip}     \fi
\ifx \showISSN     \undefined \def \showISSN      #1{\unskip}     \fi
\ifx \showLCCN     \undefined \def \showLCCN      #1{\unskip}     \fi
\ifx \shownote     \undefined \def \shownote      #1{#1}          \fi
\ifx \showarticletitle \undefined \def \showarticletitle #1{#1}   \fi
\ifx \showURL      \undefined \def \showURL       {\relax}        \fi
\providecommand\bibfield[2]{#2}
\providecommand\bibinfo[2]{#2}
\providecommand\natexlab[1]{#1}
\providecommand\showeprint[2][]{arXiv:#2}

\bibitem[Akagi et~al\mbox{.}(2019)]%
        {Akagi19HighLevelUAVGestures}
\bibfield{author}{\bibinfo{person}{John Akagi}, \bibinfo{person}{Brady Moon}, \bibinfo{person}{Xingguang Chen}, {and} \bibinfo{person}{Cameron~K. Peterson}.} \bibinfo{year}{2019}\natexlab{}.
\newblock \showarticletitle{Gesture Commands for Controlling High-Level UAV Behavior}. In \bibinfo{booktitle}{\emph{Proceedings of the International Conference on Unmanned Aircraft Systems (ICUAS)}} \emph{(\bibinfo{series}{ICUAS '19})}. \bibinfo{publisher}{IEEE}, \bibinfo{pages}{1023--1030}.
\newblock
\href{https://doi.org/10.1109/ICUAS.2019.8797743}{doi:\nolinkurl{10.1109/ICUAS.2019.8797743}}


\bibitem[DelPreto and Rus(2020)]%
        {DelPreto20PlugAndPlay}
\bibfield{author}{\bibinfo{person}{Joseph DelPreto} {and} \bibinfo{person}{Daniela Rus}.} \bibinfo{year}{2020}\natexlab{}.
\newblock \showarticletitle{Plug-and-Play Gesture Control Using Muscle and Motion Sensors}. In \bibinfo{booktitle}{\emph{Proceedings of the ACM/IEEE International Conference on Human-Robot Interaction}} \emph{(\bibinfo{series}{HRI '20})}. \bibinfo{publisher}{ACM}, \bibinfo{pages}{439--448}.
\newblock
\href{https://doi.org/10.1145/3319502.3374823}{doi:\nolinkurl{10.1145/3319502.3374823}}


\bibitem[Gromov et~al\mbox{.}(2019)]%
        {Gromov19PointingIMU}
\bibfield{author}{\bibinfo{person}{Boris Gromov}, \bibinfo{person}{Gabriele Abbate}, \bibinfo{person}{Luca~M. Gambardella}, {and} \bibinfo{person}{Alessandro Giusti}.} \bibinfo{year}{2019}\natexlab{}.
\newblock \showarticletitle{Proximity Human-Robot Interaction Using Pointing Gestures and a Wrist-mounted IMU}. In \bibinfo{booktitle}{\emph{Proceedings of the IEEE International Conference on Robotics and Automation (ICRA)}} \emph{(\bibinfo{series}{ICRA '19})}. \bibinfo{publisher}{IEEE}, \bibinfo{pages}{8084--8091}.
\newblock
\href{https://doi.org/10.1109/ICRA.2019.8794399}{doi:\nolinkurl{10.1109/ICRA.2019.8794399}}


\bibitem[Harrison et~al\mbox{.}(2025)]%
        {Harrison25IMUSmartGlove}
\bibfield{author}{\bibinfo{person}{Amy Harrison}, \bibinfo{person}{Andrea Jester}, \bibinfo{person}{Surej Mouli}, \bibinfo{person}{Antonio Fratini}, {and} \bibinfo{person}{Ali Jabran}.} \bibinfo{year}{2025}\natexlab{}.
\newblock \showarticletitle{Systematic Evaluation of IMU Sensors for Application in Smart Glove System for Remote Monitoring of Hand Differences}.
\newblock \bibinfo{journal}{\emph{Sensors}} \bibinfo{volume}{25}, \bibinfo{number}{1}, Article \bibinfo{articleno}{2} (\bibinfo{year}{2025}).
\newblock
\href{https://doi.org/10.3390/s25010002}{doi:\nolinkurl{10.3390/s25010002}}


\bibitem[Jaramillo et~al\mbox{.}(2022)]%
        {Jaramillo22}
\bibfield{author}{\bibinfo{person}{Ismael~Espinoza Jaramillo}, \bibinfo{person}{Jin~Gyun Jeong}, \bibinfo{person}{Patricio~Rivera Lopez}, \bibinfo{person}{Choong-Ho Lee}, \bibinfo{person}{Do-Yeon Kang}, \bibinfo{person}{Tae-Jun Ha}, \bibinfo{person}{Ji-Heon Oh}, \bibinfo{person}{Hwanseok Jung}, \bibinfo{person}{Jin~Hyuk Lee}, \bibinfo{person}{Won~Hee Lee}, {and} \bibinfo{person}{Tae-Seong Kim}.} \bibinfo{year}{2022}\natexlab{}.
\newblock \showarticletitle{Real-Time Human Activity Recognition with IMU and Encoder Sensors in Wearable Exoskeleton Robot via Deep Learning Networks}.
\newblock \bibinfo{journal}{\emph{Sensors}} \bibinfo{volume}{22}, \bibinfo{number}{24}, Article \bibinfo{articleno}{9690} (\bibinfo{year}{2022}).
\newblock
\href{https://doi.org/10.3390/s22249690}{doi:\nolinkurl{10.3390/s22249690}}


\bibitem[Labazanova et~al\mbox{.}(2019)]%
        {Labazanova19SwarmGlove}
\bibfield{author}{\bibinfo{person}{Luiza Labazanova}, \bibinfo{person}{Akerke Tleugazy}, \bibinfo{person}{Evgeny Tsykunov}, {and} \bibinfo{person}{Dzmitry Tsetserukou}.} \bibinfo{year}{2019}\natexlab{}.
\newblock \showarticletitle{{SwarmGlove}: A Wearable Tactile Device for Navigation of Swarm of Drones in {VR} Environment}.
\newblock In \bibinfo{booktitle}{\emph{Haptic Interaction}}, \bibfield{editor}{\bibinfo{person}{Hiroyuki Kajimoto}, \bibinfo{person}{Dongjun Lee}, \bibinfo{person}{Sang-Youn Kim}, \bibinfo{person}{Masashi Konyo}, {and} \bibinfo{person}{Ki-Uk Kyung}} (Eds.). \bibinfo{series}{Lecture Notes in Electrical Engineering}, Vol.~\bibinfo{volume}{535}. \bibinfo{publisher}{Springer}, \bibinfo{address}{Singapore}, \bibinfo{pages}{304--309}.
\newblock
\href{https://doi.org/10.1007/978-981-13-3194-7_67}{doi:\nolinkurl{10.1007/978-981-13-3194-7_67}}


\bibitem[Lin et~al\mbox{.}(2018)]%
        {Lin18}
\bibfield{author}{\bibinfo{person}{Bor-Shing Lin}, \bibinfo{person}{I-Jung Lee}, \bibinfo{person}{Shu-Yu Yang}, \bibinfo{person}{Yi-Chiang Lo}, \bibinfo{person}{Junghsi Lee}, {and} \bibinfo{person}{Jean-Lon Chen}.} \bibinfo{year}{2018}\natexlab{}.
\newblock \showarticletitle{Design of an Inertial-Sensor-Based Data Glove for Hand Function Evaluation}.
\newblock \bibinfo{journal}{\emph{Sensors}} \bibinfo{volume}{18}, \bibinfo{number}{5}, Article \bibinfo{articleno}{1545} (\bibinfo{year}{2018}).
\newblock
\href{https://doi.org/10.3390/s18051545}{doi:\nolinkurl{10.3390/s18051545}}


\bibitem[Madgwick(2010)]%
        {Madgwick10}
\bibfield{author}{\bibinfo{person}{Sebastian O.~H. Madgwick}.} \bibinfo{year}{2010}\natexlab{}.
\newblock \bibinfo{booktitle}{\emph{An Efficient Orientation Filter for Inertial and Inertial/Magnetic Sensor Arrays}}.
\newblock \bibinfo{type}{{T}echnical {R}eport}. \bibinfo{institution}{University of Bristol and x-io Technologies}.
\newblock


\bibitem[Pfeiffer and Scaramuzza(2021)]%
        {Pfeiffer21DroneRacing}
\bibfield{author}{\bibinfo{person}{Christian Pfeiffer} {and} \bibinfo{person}{Davide Scaramuzza}.} \bibinfo{year}{2021}\natexlab{}.
\newblock \showarticletitle{Human-Piloted Drone Racing: Visual Processing and Control}.
\newblock \bibinfo{journal}{\emph{IEEE Robotics and Automation Letters}} \bibinfo{volume}{6}, \bibinfo{number}{2} (\bibinfo{year}{2021}), \bibinfo{pages}{3467--3474}.
\newblock
\href{https://doi.org/10.1109/LRA.2021.3064282}{doi:\nolinkurl{10.1109/LRA.2021.3064282}}


\bibitem[Serpiva et~al\mbox{.}(2024)]%
        {Serpiva24OmniRace}
\bibfield{author}{\bibinfo{person}{Valerii Serpiva}, \bibinfo{person}{Aleksey Fedoseev}, \bibinfo{person}{Sausar Karaf}, \bibinfo{person}{Ali~Alridha Abdulkarim}, {and} \bibinfo{person}{Dzmitry Tsetserukou}.} \bibinfo{year}{2024}\natexlab{}.
\newblock \showarticletitle{OmniRace: 6D Hand Pose Estimation for Intuitive Guidance of Racing Drone}. In \bibinfo{booktitle}{\emph{2024 IEEE/RSJ International Conference on Intelligent Robots and Systems (IROS)}}. \bibinfo{pages}{2508--2513}.
\newblock
\href{https://doi.org/10.1109/IROS58592.2024.10801907}{doi:\nolinkurl{10.1109/IROS58592.2024.10801907}}


\bibitem[Serpiva et~al\mbox{.}(2021)]%
        {Serpiva21Swarmpaint}
\bibfield{author}{\bibinfo{person}{Valerii Serpiva}, \bibinfo{person}{Ekaterina Karmanova}, \bibinfo{person}{Aleksey Fedoseev}, \bibinfo{person}{Stepan Perminov}, {and} \bibinfo{person}{Dzmitry Tsetserukou}.} \bibinfo{year}{2021}\natexlab{}.
\newblock \showarticletitle{Swarmpaint: Human-swarm Interaction for Trajectory Generation and Formation Control by DNN-based Gesture Interface}. In \bibinfo{booktitle}{\emph{Proceedings of the International Conference on Unmanned Aircraft Systems (ICUAS)}} \emph{(\bibinfo{series}{ICUAS '21})}. \bibinfo{publisher}{IEEE}, \bibinfo{pages}{1055--1062}.
\newblock
\href{https://doi.org/10.1109/ICUAS51884.2021.9476795}{doi:\nolinkurl{10.1109/ICUAS51884.2021.9476795}}


\bibitem[Tezza and Andujar(2019)]%
        {Tezza19HumanDroneSurvey}
\bibfield{author}{\bibinfo{person}{Dante Tezza} {and} \bibinfo{person}{Marvin Andujar}.} \bibinfo{year}{2019}\natexlab{}.
\newblock \showarticletitle{The State-of-the-Art of Human--Drone Interaction: A Survey}.
\newblock \bibinfo{journal}{\emph{IEEE Access}}  \bibinfo{volume}{7} (\bibinfo{year}{2019}), \bibinfo{pages}{167438--167454}.
\newblock
\href{https://doi.org/10.1109/ACCESS.2019.2953900}{doi:\nolinkurl{10.1109/ACCESS.2019.2953900}}


\bibitem[Tokmurziyev et~al\mbox{.}(2024)]%
        {GazeRace}
\bibfield{author}{\bibinfo{person}{Issatay Tokmurziyev}, \bibinfo{person}{Valerii Serpiva}, \bibinfo{person}{Aleksey Fedoseev}, \bibinfo{person}{Miguel~Altamirano Cabrera}, {and} \bibinfo{person}{Dzmitry Tsetserukou}.} \bibinfo{year}{2024}\natexlab{}.
\newblock \showarticletitle{GazeRace: Revolutionizing Remote Piloting with Eye-Gaze Control}. In \bibinfo{booktitle}{\emph{Proc IEEE Int. Conf. on Systems, Man, and Cybernetics (SMC)}}. \bibinfo{pages}{410--415}.
\newblock
\href{https://doi.org/10.1109/SMC54092.2024.10831987}{doi:\nolinkurl{10.1109/SMC54092.2024.10831987}}


\bibitem[Trinitatova et~al\mbox{.}(2025)]%
        {Trinitatova25Handheld}
\bibfield{author}{\bibinfo{person}{Daria Trinitatova}, \bibinfo{person}{Sofia Shevelo}, {and} \bibinfo{person}{Dzmitry Tsetserukou}.} \bibinfo{year}{2025}\natexlab{}.
\newblock \showarticletitle{Towards Intuitive Drone Operation Using a Handheld Motion Controller}. In \bibinfo{booktitle}{\emph{2025 20th ACM/IEEE International Conference on Human-Robot Interaction (HRI)}}. \bibinfo{pages}{1690--1694}.
\newblock
\href{https://doi.org/10.1109/HRI61500.2025.10974242}{doi:\nolinkurl{10.1109/HRI61500.2025.10974242}}


\bibitem[Yang et~al\mbox{.}(2020)]%
        {Yang20}
\bibfield{author}{\bibinfo{person}{Zhicheng Yang}, \bibinfo{person}{Bert-Jan~F. van Beijnum}, \bibinfo{person}{Bin Li}, \bibinfo{person}{Shenggang Yan}, {and} \bibinfo{person}{Peter~H. Veltink}.} \bibinfo{year}{2020}\natexlab{}.
\newblock \showarticletitle{Estimation of Relative Hand-Finger Orientation Using a Small IMU Configuration}.
\newblock \bibinfo{journal}{\emph{Sensors}} \bibinfo{volume}{20}, \bibinfo{number}{14}, Article \bibinfo{articleno}{4008} (\bibinfo{year}{2020}).
\newblock
\href{https://doi.org/10.3390/s20144008}{doi:\nolinkurl{10.3390/s20144008}}


\bibitem[Yashin et~al\mbox{.}(2019)]%
        {Yashin19AeroVR}
\bibfield{author}{\bibinfo{person}{Grigoriy~A. Yashin}, \bibinfo{person}{Daria Trinitatova}, \bibinfo{person}{Ruslan~T. Agishev}, \bibinfo{person}{Roman Ibrahimov}, {and} \bibinfo{person}{Dzmitry Tsetserukou}.} \bibinfo{year}{2019}\natexlab{}.
\newblock \showarticletitle{AeroVr: Virtual Reality-based Teleoperation with Tactile Feedback for Aerial Manipulation}. In \bibinfo{booktitle}{\emph{2019 19th International Conference on Advanced Robotics (ICAR)}}. \bibinfo{pages}{767--772}.
\newblock
\href{https://doi.org/10.1109/ICAR46387.2019.8981574}{doi:\nolinkurl{10.1109/ICAR46387.2019.8981574}}


\end{thebibliography}
\end{document}